\documentclass[journal]{IEEEtran}

\usepackage{makecell}
\usepackage[utf8]{inputenc}
\usepackage{graphicx}
\usepackage{amsmath,amsfonts,amssymb}
\usepackage{cite}
\usepackage{booktabs}
\usepackage{multirow}
\usepackage{url}
\usepackage{xcolor}
\usepackage{subfig}
\usepackage[colorlinks=true, linkcolor=blue, urlcolor=blue, citecolor=blue]{hyperref}
\usepackage[compatibility=false]{caption}

\title{A Transformer-Based Conditional GAN with Multiple Instance Learning for UAV Signal Detection and Classification}

\author{
  \IEEEauthorblockN{Haochen Liu$^{*}$, Jia Bi$^{*}$, Xiaomin Wang, Xin Yang, Ling Wang}
\thanks{$^{*}$Haochen Liu and Jia Bi contributed equally to this work.}
\thanks{J. Bi is with Scientific Computing Department, STFC, Rutherford Appleton Laboratory, Didcot OX11 0QX, United Kingdom. (e-mail: Jia.Bi@stfc.ac.uk).}
\thanks{H. Liu, X. Wang, X. Yang and L.Wang are with the School of Electronics and Information, Northwestern Polytechnical University, Xi'an 710019, China (email:\{haochenliu, xiaominwang,xinyang,lingwang\}@nwpu.edu.cn).}}

\begin{document}
\maketitle

\begin{abstract}
Unmanned Aerial Vehicles (UAVs) have gained significant traction across a multitude of domains including surveillance, logistics, agriculture, disaster management, and military operations. 
To maintain effective UAV operations, it is crucial to accurately detect and classify their flight states, such as hovering, cruising, ascending, or transitioning between these modes.
Conventional Time Series Classification (TSC) methods often lack the robustness and generalization required for dynamic and heterogeneous UAV environments, while state-of-the-art models such as Transformers and Long Short Term Memory (LSTM)-based architectures typically demand large datasets and incur high computational costs, especially when handling high-dimensional data streams. 
In this paper, we propose a novel Transformer-based Generative Adversarial Network (GAN) framework integrated with Multiple Instance Locally Explainable Learning (MILET) to tackle these challenges in UAV flight state classification. 
The proposed method leverages a Transformer encoder to capture long-range temporal dependencies and complex dynamics from UAV telemetry, while a GAN is employed to augment limited training datasets with realistic synthetic samples. 
MIL principles are integrated to effectively handle high-dimensional inputs by focusing attention on the most discriminative segments, thereby mitigating noise and computational overhead.
The experimental results demonstrate the superior performance of the proposed method, achieving the highest classification accuracy of 96.5\% on the \textit{DroneDetect} dataset and 98.6\% on the \textit{DroneRF} dataset compared to other state-of-the-art methods.
The framework excels in computational efficiency and robust generalization across various UAV platforms and flight states.
These results underscore its strong potential for real-time deployment in resource-constrained environments, offering a scalable and resilient solution for UAV flight state classification in dynamic and complex operational scenarios.

\end{abstract}

\begin{IEEEkeywords}
UAV, Flight State Classification, Multiple Instance Learning, Transformer, GAN, Time Series Classification
\end{IEEEkeywords}

\section{Introduction}

Unmanned Aerial Vehicles (UAVs) have revolutionized numerous industries, enabling a wide array of applications, including surveillance, logistics, environmental monitoring, precision agriculture, disaster relief operations, and military missions \cite{austin2010unmanned, guerrero2007multiple, yanmaz2018drone}. They offer capabilities such as hovering at fixed points for surveillance \cite{chandler2004uav}, cruising over large areas for reconnaissance \cite{waharte2010supporting}, ascending or descending to specific altitudes for payload delivery \cite{khaleghi2018uav}, and smoothly transitioning between states to adapt to changing flight conditions \cite{nex2014uav}. As these platforms continue to mature, it becomes paramount to ensure their safe, efficient, and effective operation \cite{rutherford2007framework}.

A core challenge in UAV flight state classification lies in achieving robust and generalizable performance under real-world constraints. Accurate real-time detection of whether a UAV is hovering, cruising, ascending, or transitioning between states is critical to ensure safe autonomous navigation, effective collision avoidance, and reliable threat detection \cite{Zhang2019,Rana2020}. However, several significant issues impede the development of efficient classification systems. TSC methods often lack the robustness necessary to generalize across diverse operational environments and varying UAV platforms, as they typically rely on handcrafted features that may not capture the underlying complexities of different flight states, leading to poor performance when applied to unseen data or novel UAV models \cite{Fawaz2019}. Furthermore, UAV telemetry data is inherently high-dimensional, encompassing multiple correlated parameters such as velocity, orientation, altitude, and position \cite{Marconato2021,Valverde-Colmeiro2023}, which require sophisticated algorithms to process and analyze effectively. This high dimensionality significantly increases computational costs, making it challenging to implement real-time classification on resource-constrained UAV platforms \cite{Zerveas2021}. Furthermore, the scarcity of representative labeled data, driven by the high costs associated with extensive data collection and annotation \cite{Delmerico2019}, exacerbates these challenges by limiting the ability of deep learning models to learn generalized and robust representations, often resulting in models that are prone to overfitting and exhibit diminished performance in diverse and dynamic environments. Addressing these intertwined issues requires innovative approaches that enhance generalization and robustness while efficiently managing high-dimensional data and mitigating the effects of limited training resources.

To address these intertwined challenges, we propose an innovative framework that synergistically combines Transformer-based feature extraction \cite{Vaswani2017,Zerveas2021}, Generative Adversarial Networks (GANs) for data augmentation \cite{Goodfellow2014,Mirza2014}, and Multiple Instance Learning for Locally Explainable Time Series Classification (MILET) \cite{early2024inherently}, termed \textit{Trans\_GAN\_Milet}. This integration aims to enhance the generalization and robustness of UAV flight state classification models while efficiently managing high-dimensional data and mitigating the limitations posed by scarce training resources. Specifically, our Transformer-based module captures long-range temporal dependencies and complex interactions within high-dimensional telemetry data, enabling the model to generalize across various UAV platforms and operational environments \cite{Vaswani2017}. The GAN component generates realistic synthetic telemetry samples, augmenting the limited training data and improving the robustness of the model against overfitting \cite{Goodfellow2014}. Meanwhile, MILET employs Multiple Instance Learning to focus the model's attention on the most informative segments of flight data, thereby reducing computational overhead and enhancing classification accuracy by emphasizing critical instances that significantly contribute to flight state determination \cite{Ilse2018}. Furthermore, the locally explainable nature of MILET provides interpretability to the classification results, facilitating better understanding and trust in the model predictions. Through this comprehensive framework, \textit{Trans\_GAN\_MILET} effectively addresses the core challenges of limited generalization, insufficient robustness, high-dimensional data processing, and data scarcity in UAV flight state classification.

Our main contributions can be summarized as follows:
The main contributions of this paper are as follows:
\begin{itemize}
    \item \textbf{Unified Hybrid Framework:} We design an end-to-end architecture that integrates Transformer-based temporal modeling, GAN-based data augmentation, and locally explainable MIL, achieving robust UAV flight state classification under high-dimensional and limited data conditions.
    \item \textbf{Efficient Feature Extraction:} The Transformer encoder with MIL pooling efficiently captures long-range dependencies and highlights salient signal segments, boosting accuracy and interpretability.
    \item \textbf{Data-Efficient Learning:} The conditional GAN module generates high-fidelity synthetic telemetry data, expanding training sets and reducing overfitting in small-sample scenarios.
    \item \textbf{Comprehensive Empirical Validation:} Extensive experiments on public and proprietary UAV datasets demonstrate that our approach outperforms state-of-the-art baselines in accuracy, efficiency, and generalization, and is suitable for real-time deployment.
\end{itemize}

\section{Related Work}

Research in UAV flight state classification encompasses a wide range of methodologies, evolving from traditional signal processing and classical machine learning techniques to advanced deep learning architectures. This progression reflects the growing complexity and demands of accurately detecting and classifying UAV flight states in diverse and dynamic environments.

\subsection{UAV Flight State Classification Approaches}

Early studies in UAV flight state classification primarily relied on engineered features extracted from sensor data, such as accelerometers, gyroscopes, and RF signals, in conjunction with classical classifiers such as k-Nearest Neighbors (k-NN) and Support Vector Machines (SVM) \cite{Zhang2019}. While these methods offered interpretability and simplicity, they often lacked scalability and adaptability, particularly in environments with rapidly changing flight conditions. To overcome these limitations, more recent research has leveraged advanced deep learning architectures, including Convolutional Neural Networks (CNNs), Long Short-Term Memory (LSTM) networks, and hybrid models that combine both \cite{Rana2020, Fawaz2019}. These approaches significantly improved classification accuracy by automatically learning complex temporal and local spatial dependencies directly from raw telemetry data. However, despite these advancements, CNN and LSTM-based models exhibit notable drawbacks. Specifically, they struggle to capture global relationships within time series data, which are crucial for understanding long-term dependencies and contextual information throughout the entire flight sequence. Additionally, these models often require sequential processing, hindering their ability to be effectively parallelized. This sequential nature not only increases computational latency but also makes real-time detection challenging, particularly on resource-constrained UAV platforms where processing power and memory are limited. For example, the integration of multi-channel CNNs (MC-CNN) with Feature Engineering Generators (FEG) has emerged as a promising approach to further enhance UAV flight state classification \cite{e23121678}. However, the reliance on extensive feature engineering limits the model's adaptability to different UAV types and varying operational environments, as the engineered features may not generalize well across diverse conditions. Furthermore, the increased computational complexity associated with processing multiple frequency channels results in longer training and inference times, rendering the approach less suitable for real-time applications on UAV platforms with limited computational resources.

\begin{figure*}[t]
    \centering
    \includegraphics[width=0.8\textwidth]{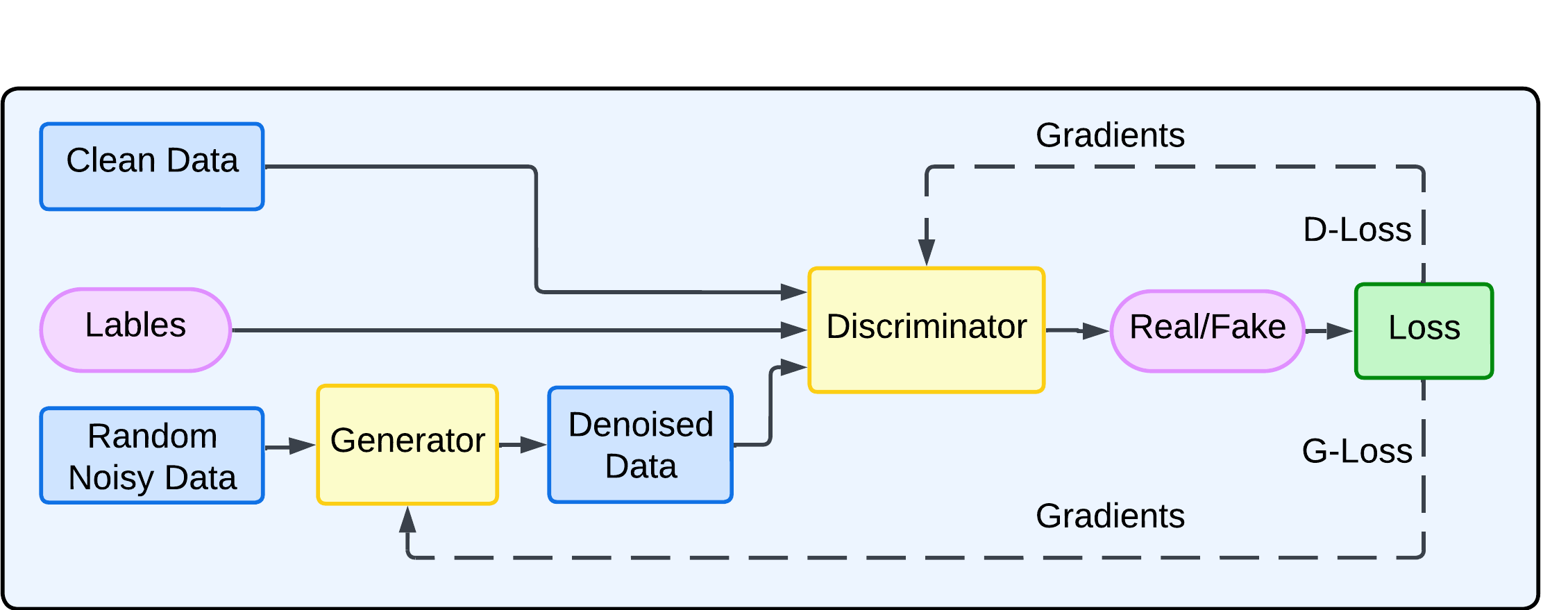}
    \caption{Proposed framework integrating a Transformer-based encoder, a GAN for data augmentation, and MIL for effective UAV flight state classification.}
    \label{framework}
\end{figure*}
\subsection{Time Series Classification and Feature Extraction for UAV Applications}

TSC has undergone significant advancements, evolving from traditional techniques such as Dynamic Time Warping (DTW) and dictionary-based methods to the adoption of sophisticated deep learning architectures \cite{Fawaz2019}. Early deep learning models like Long Short-Term Memory (LSTM) and Gated Recurrent Units (GRUs) enhanced classification performance by effectively capturing complex temporal dependencies inherent in time series data. However, these models are often hindered by sequential processing overhead and a limited capacity to efficiently model very long-range dependencies, which can restrict their performance in applications that require real-time analysis and interpretation \cite{Rana2020}. To address these limitations, Transformer-based architectures \cite{Vaswani2017}, originally developed for Natural Language Processing (NLP), have been adapted for TSC tasks. Transformers introduce an attention-based mechanism that decouples processing time from sequence length, enabling parallel processing of data and more efficient handling of long-range dependencies. This makes them particularly suitable for managing complex and high-dimensional telemetry signals generated by UAVs, which require the ability to capture both local and global patterns across multiple flight parameters. Transformers have shown exceptional promise in various TSC applications, including forecasting, anomaly detection, and classification, utilizing multihead self-attention to learn intricate contextual relationships within the data \cite{Zerveas2021}. In UAV-related tasks, Transformers excel at capturing global correlations across different flight parameters without relying solely on recurrent structures, resulting in more accurate and robust classification of diverse flight states. Despite these advantages, the implementation of Transformers in UAV scenarios is often challenged by two primary factors: high computational costs, especially for high-dimensional datasets, and the requirement for large and diverse training datasets—a condition frequently unmet due to the high costs and logistical complexities associated with collecting and annotating extensive flight telemetry data.

\begin{figure*}[t]
    \centering
    \includegraphics[width=1\textwidth]{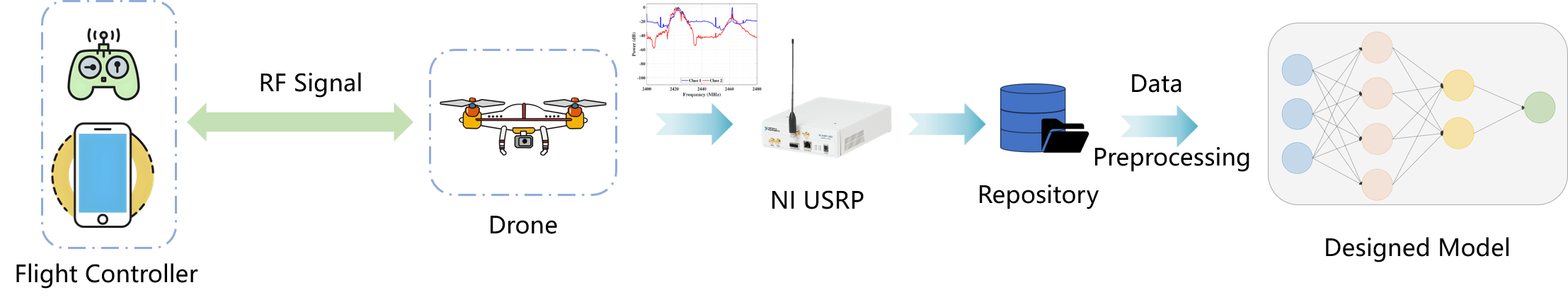}
    \caption{System Model Block Diagram}
    \label{system_model1}
\end{figure*}

\subsection{Multiple Instance Learning}

Multiple Instance Learning (MIL) \cite{Dietterich1997} is a machine learning framework where training data is organized into \textit{bags} of instances, with labels assigned to entire bags rather than individual instances. This approach is particularly beneficial when instance-level labeling is impractical or infeasible. In UAV telemetry analysis, not every segment of a flight log contains significant information for flight state classification; only specific segments hold the critical cues necessary for accurate state determination. Traditionally, MIL is treated as a binary classification problem: a bag is labeled positive if at least one instance within it is positive, and negative otherwise \cite{Dietterich1997}. However, in designing MILLET—a general and widely applicable TSC approach—we relax this assumption to focus on capturing temporal relationships, meaning the order of instances within bags matters. This flexibility allows us to incorporate positional encodings into our MILLET methods, enhancing the model's ability to recognize and leverage sequential patterns in the data.

Thus, by integrating MIL with advanced techniques like Transformers and Generative Adversarial Networks (GANs), our work addresses these limitations. MIL enables the model to focus on the most informative flight log segments, efficiently capturing both local and global relationships in high-dimensional datasets. Additionally, GANs generate synthetic data to augment the true data, thereby improving model generalization and robustness while reducing computational costs—crucial for real-time applications on resource-constrained UAV platforms.

\section{System Model and Dataset Discription}
Based on the Unmanned Aerial Vehicle (UAV) flight scenario, this section builds a UAV flight status recognition system, including UAV communication module, signal interception module and processing recognition module. At the same time, this section provides corresponding descriptions and explanations of the dataset composed of intercepted signals, laying the foundation for the input preprocessing of the subsequent processing recognition module.

\subsection{System Model}

The UAV flight status recognition system is shown in Fig.~\ref{system_model1}, which mainly includes the flight controller, drone, RF sensing module integrated with USRP, data processing module and a neural network.

Flight controllers mainly include various remote controllers, mobile phones and applications used to control various drones. When using different control commands, different RF signals are sent, and the drone receives RF signals and performs corresponding flight actions. Different control commands can affect RF signals, and the model and communication frequency band of the drone can also have a certain impact on RF signals. The RF sensing module in USRP equipment can receive RF signals in multiple frequency bands. By combining with the LabVIEW development environment, the corresponding signal parameters can be obtained to better process theses data.

By changing control instructions, drone types, communication environments and other factors, an RF signal database can be constructed.Then a series of preprocessing operations are performed to make the dataset more suitable to the input requirements of the designed neural network.

\subsection{Dataset Discription}
The datasets used in this article are the Drone Detect dataset and the Drone RF dataset, which include RF signals generated by different types of drones in different flight modes. Acquisition of the RF signal dataset in the Repository module of system model shown in Fig~\ref{system_model1}. The two datasets will be introduced as follows.

\subsubsection{Drone Detect Dataset}

The Drone Detect dataset uses Nuand BladeRF SDR and open-source software GNU Radio to collect Unmanned Aerial System(UAS) signal segments, including four subsets of data: UAS signals with Bluetooth interference, UAS signals with Wi-Fi interference, and UAS signals with both and no interference. The parameter settings used for data collection are shown in the Table~\ref{tab:parameter_setting_1}.

\begin{table}[ht]
    \centering
    \caption{Parameter Setting in the Drone Detect Dataset}
    \label{tab:parameter_setting_1}
    \begin{tabular}{lcc}
        \toprule
        \textbf{Parameter} & \textbf{Value} \\
        \midrule
        Sampling Rate & 60Mbit/s  \\
        Bandwidth & 28MHz  \\
        Center Frequency & 2.4375GHz \\
        \bottomrule
    \end{tabular}
\end{table}

\begin{figure}[t]
    \centering
    \includegraphics[width=0.5\textwidth]{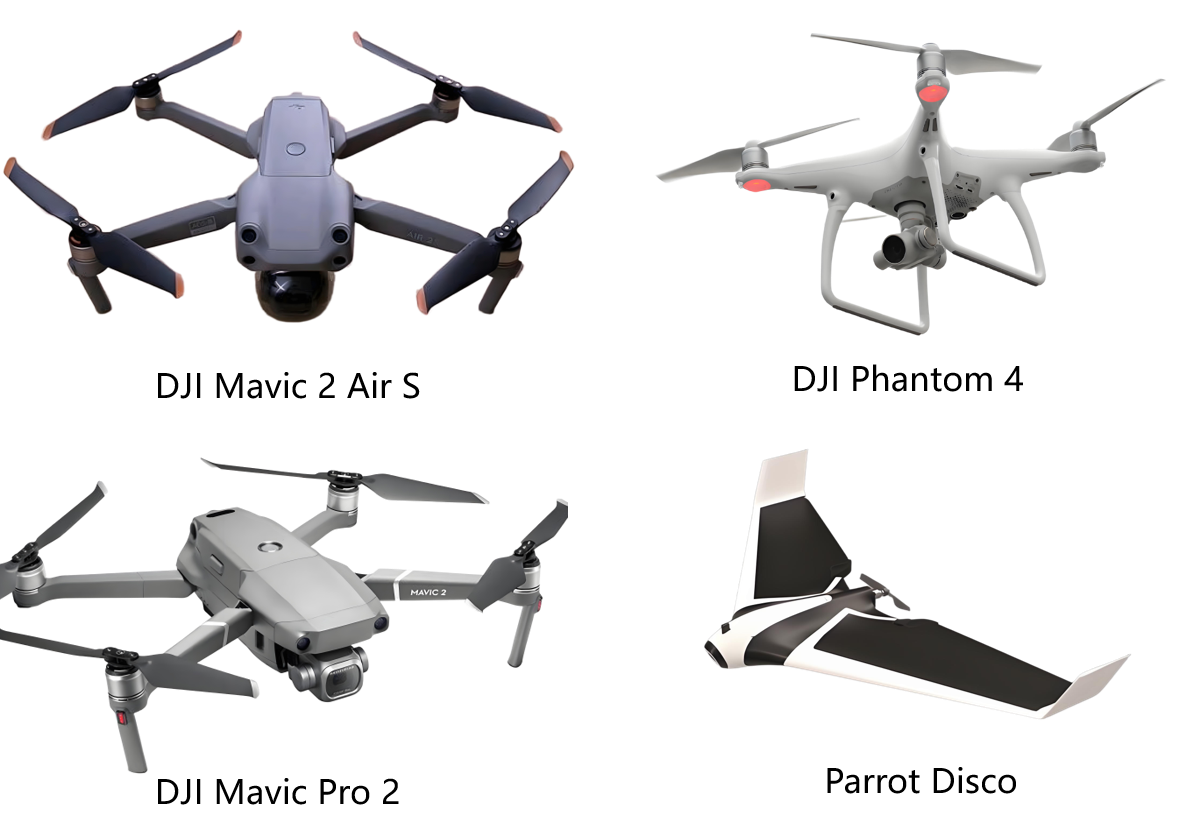}
    \caption{Some types of Drones}
    \label{UAV_picture}
\end{figure}

The dataset consists of seven different types of drones, including DJI Mavic 2 Air S, DJI Mavic Pro, DJI Mavic Pro 2, DJI Inspire 2, DJI Mavic Mini, DJI Phantom 4 and Parrot Disco. There are three flight modes: Switched on(ON), Hovering(HO), and Flying(FY). Therefore, the dataset consists of 21 types of signals. Fig.~\ref{UAV_picture} shows some drones. It can be seen that the construction and purpose of different models of drones are different, resulting in different flight states under the same flight mode.

\subsubsection{Drone RF Dataset}

Drone RF dataset was proposed by Mohammad F. al Sa'd et al. in 2019. This dataset considers the absence and presence of drones in RF backgrounds. At the same time, it also contains RF signals from different drones in different flight modes. This dataset covers a wide range and can simulate commonly used flight scenarios, which has important reference significance for the study of UAV flight state recognition.

The Drone RF dataset includes three different drones (Parrot Bebop, Parrot AR Drone, DJI Phantom 3) and four flight modes (off, on and connected, hover, flight and video recording). Table~\ref{tab:specification_drone} lists the main characteristics of these drones.

\begin{table}[ht]
    \centering
    \caption{Specifications of drones under analysis}
    \label{tab:specification_drone}
    \begin{tabular}{lcccc}
        \toprule
        \textbf{Drone} & \textbf{Parrot Bebop} & \textbf{Parrot AR Drone} & \textbf{DJI Phantom 3} \\
        \midrule
        Dimensions(cm) & 38×33×3.6 & 61×61×12.7 & 52×49×29 \\
        Weight(g) & 400 & 420 & 1216 \\
        \makecell{Battery capacity \\ (mAh)} & 1200 & 1000 & 4480 \\
        Max.range(m) & 250 & 50 &1000 \\
        Connectivity & \makecell{WiFi(2.4 GHz \\ and 5 GHz)} & WiFi(2.4 GHz) & \makecell{WiFi(2.4 GHz \\ -2.483 GHz) \\ + RF(5.725 GHz \\ -5.825 GHz)} \\
        \bottomrule
    \end{tabular}
\end{table}

The experiment intercepts the RF signals exchanged between the drone and the remote controller through the RF sensing module in NI-USRP. At the same time, LabVIEW is used to select appropriate parameters to obtain, process and store RF data segments. The parameter settings for this dataset are shown in Table~\ref{tab:parameter_setting_2}:

\begin{table}[ht]
    \centering
    \caption{Parameter Setting in the Drone RF Dataset}
    \label{tab:parameter_setting_2}
    \begin{tabular}{lcc}
        \toprule
        \textbf{Parameter} & \textbf{Value} \\
        \midrule
        Sampling Rate & 40Mbit/s  \\
        Bandwidth & 40MHz  \\
        Center Frequency & 2.422GHz \\
        \bottomrule
    \end{tabular}
\end{table}

\begin{figure}[!htb]
    \centering
    \subfloat[Comparison between RF signal emitted by Drone hovering and RF background]{\label{RF background and drone present}
		\includegraphics[scale=0.4]{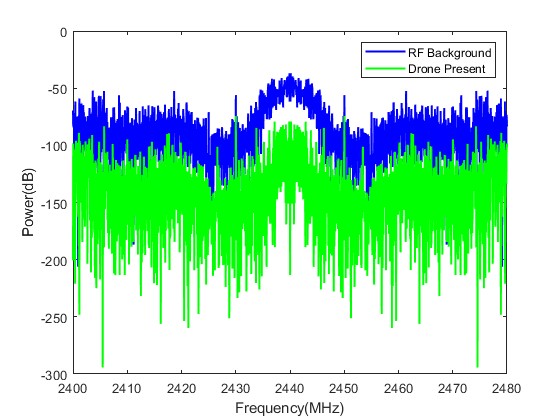}}\\
    \subfloat[Comparison of three Drones in hovering state]{\label{three drone comparision}
		\includegraphics[scale=0.4]{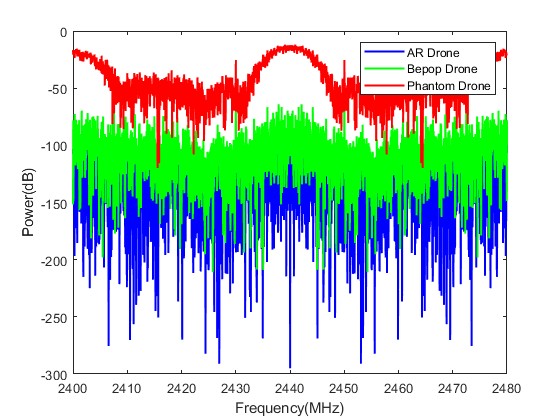}}\\
    \subfloat[Comparison of RF signals of AR Drone in four modes]{\label{four mode comparision}
		\includegraphics[scale=0.4]{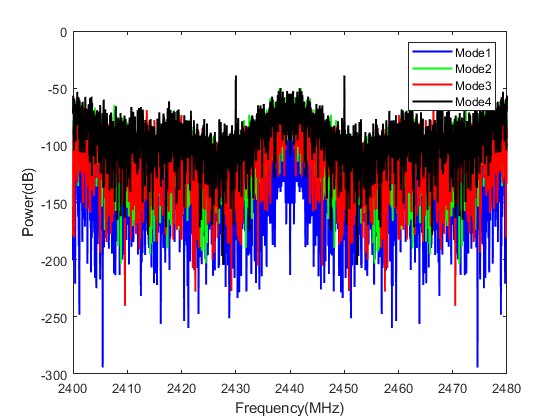}}\\
    \caption{Drone RF Dataset Analysis}
    \label{Drone RF Analysis}
\end{figure}

Fig.~\ref{Drone RF Analysis} compares the RF signal spectra under different conditions in the Drone RF dataset, with 2.44GHz as the center frequency and 2048 FFT points. Fig.~\ref{RF background and drone present} shows the comparison of RF signal spectra between RF background only and AR Drone hovering state; Fig.~\ref{three drone comparision} shows a comparison of RF signal spectra for AR Drone, Bepop Drone, and Phantom Drone; Fig.~\ref{four mode comparision} shows the RF signal spectrum comparison of AR Drone in four modes: off, on and connected, hovering, flying, and video recording. Therefore, we can conclude that different types of RF signals in the dataset have significant differences under different drones and flight modes. The flight status of drones can be recognized based on RF signals.

In UAV communication systems, two primary types of RF signals are exchanged between UAVs and their controllers:
\begin{enumerate}
    \item \textit{Uplink Signals}: These carry control commands from the ground controller to the UAV, enabling navigation and operational instructions.
    \item \textit{Downlink Signals}: These transmit telemetry data, including positional and status information, as well as video feeds from the UAV back to the controller.
\end{enumerate}

Each UAV model generates unique RF signatures influenced by specific circuitry designs and modulation schemes employed by different manufacturers. Predominantly operating within the 2.4 GHz and 5.8 GHz frequency bands, these signals are critical for distinguishing between various UAV types and their operational states.

RF signals were captured using advanced Software-Defined Radios (SDRs) such as the Nuand BladeRF \cite{nuand2019bladerf}, which are equipped with high-gain antennas and multi-channel receivers. To ensure signal fidelity, sampling rates were set above the Nyquist rate, effectively preventing aliasing. The operational bandwidth was strategically divided into low-frequency and high-frequency segments, with each segment recorded by dedicated SDR devices. This approach overcomes bandwidth limitations and ensures comprehensive coverage of all relevant frequency components necessary for subsequent analysis.

\subsubsection{Noise and Interference in RF Signals}

RF signal integrity is often compromised by various sources of noise and interference, which pose significant challenges for accurate UAV classification, which thermal noise, environmental noise, external interference, multipath effects, and intentional jamming or Spoofing. These factors collectively degrade the Signal-to-Interference-plus-Noise Ratio (SINR), resulting in captured signals that are a mixture of UAV-generated transmissions and ambient noise. This degradation complicates the extraction of meaningful features necessary for accurate UAV flight state classification.

\subsubsection{RF Signals in the Frequency Domain}

Transforming RF signals from the time domain to the frequency domain offers several advantages that are crucial for effective classification:
\begin{enumerate}
    \item \textbf{Data Reduction}: Frequency domain representations typically require less data to capture essential signal characteristics, thereby reducing computational overhead.
    \item \textbf{Noise Mitigation}: Analyzing signals in the frequency domain helps isolate and minimize noise-dominated segments, enhancing feature quality.
    \item \textbf{Bandwidth Management}: Frequency domain processing allows for the seamless concatenation and analysis of signals across different bandwidth-limited segments, addressing challenges related to bandwidth constraints.
\end{enumerate}

In this study, the Fourier Transform was employed to convert time-domain RF signals into the frequency domain. This transformation facilitates the identification of critical spectral features, such as harmonics, frequency peaks, and power distributions, which are instrumental in distinguishing between different UAV flight states and models. By leveraging these frequency domain characteristics, the proposed framework can more effectively capture the nuanced differences inherent in UAV-generated RF signals.

\subsubsection{Challenges of High-Dimensional Data}

UAV RF signal data is inherently high-dimensional, encompassing multiple frequency bands and numerous spectral features. High-dimensional datasets present several challenges:
\begin{itemize}
    \item \textbf{Computational Complexity}: Processing high-dimensional data requires significant computational resources, which can be a bottleneck for real-time applications, especially on resource-constrained UAV platforms.
    \item \textbf{Curse of Dimensionality}: As the number of dimensions increases, the volume of the feature space grows exponentially, making it difficult for models to generalize from limited training data \cite{bishop2006pattern}.
    \item \textbf{Feature Redundancy and Irrelevance}: High-dimensional data often contains redundant or irrelevant features that can degrade model performance and increase the risk of overfitting.
    \item \textbf{Scalability}: Models must be scalable to handle varying data sizes and complexities without compromising performance.
\end{itemize}

Addressing these challenges requires sophisticated techniques for feature extraction, dimensionality reduction, and efficient data processing to ensure that the classification model remains both accurate and computationally feasible.

\subsection{Data Preprocessing}
To ensure robust performance within our framework, we implement a comprehensive preprocessing pipeline for UAV RF signal data. This pipeline prepares clean data for the discriminator and serves as a reliable reference during adversarial training. The preprocessing pipeline encompasses the following key stages. Each of these stages targets specific aspects of the RF signal data, improving either the time-domain or frequency-domain representations to facilitate effective feature extraction and classification.

\subsubsection{\textbf{Doppler Shift Mitigation}}
Mitigating Doppler-induced frequency distortions caused by UAV dynamics—such as velocity, rotation, and other movements—is critical for accurate signal processing \cite{doppler_mitigation_ref}. The Doppler shift is mathematically defined as:
\begin{equation}
    f_d = \frac{v}{c} f_c \cos(\theta),
\end{equation}
where \( f_d \) represents the Doppler shift, \( v \) is the relative velocity between the UAV and the receiver, \( c \) is the speed of light, \( f_c \) is the carrier frequency, and \( \theta \) is the angle between the motion direction and the line of sight. These parameters are computed on the basis of various states of the UAV, including flying, hovering, and transitioning between operational modes. By compensating for the Doppler shift using this formulation, the pipeline ensures that the frequency domain data aligns with expected signal conditions. This mitigation of frequency distortions enhances the spectral integrity of the RF signals, thereby improving the accuracy and robustness of subsequent signal processing and classification tasks across diverse real-world scenarios.

\subsubsection{\textbf{Frequency Band Normalization via Filter Bank}}

RF signals consist of low-frequency and high-frequency components with distinct characteristics and power levels \cite{e23121678}. Uniform normalization can cause smaller amplitude components to be overshadowed by larger ones, obscuring their variations \cite{filter_bank_ref}. To address this, we employ a filter bank to decompose the signal into multiple frequency bands, enabling finer granularity in frequency separation. Each decomposed band undergoes independent z-score normalization, which prevents feature domination by ensuring that each frequency band maintains its unique attributes without being influenced by varying signal strengths. The normalized bands are then concatenated into a unified representation for subsequent processing. This method effectively preserves the unique spectral features of each frequency band while ensuring balanced feature scaling, making it well-suited for handling high-dimensional frequency domain RF signal data.

\subsubsection{\textbf{Temporal Segmentation via Overlapping Sliding Windows}}
To capture localized temporal features and enhance computational efficiency, we segment the time-domain signal into overlapping sliding windows. Each window captures a segment of length \( N \), defined as:
\begin{equation}
    x_{i,k} = \{x_{k}, x_{k+1}, \ldots, x_{k+N-1}\}, \quad k = 1, 2, \ldots,
\end{equation}
where \( N \) is the window length and \( k \) is the starting index. Overlapping windows ensure continuity and prevent information loss between segments. These segmented windows are directly fed into the model, leveraging the position embedding layer within the our model especially for the generator to capture sequence order and temporal dependencies. This segmentation strategy enhances the time domain data by facilitating the extraction of transient features and temporal patterns, improving computational efficiency and seamlessly integrating with the model architecture to support robust feature extraction while maintaining temporal integrity.

\section{Methodology}

\begin{figure*}[t]
    \centering
    \includegraphics[width=0.9\textwidth]{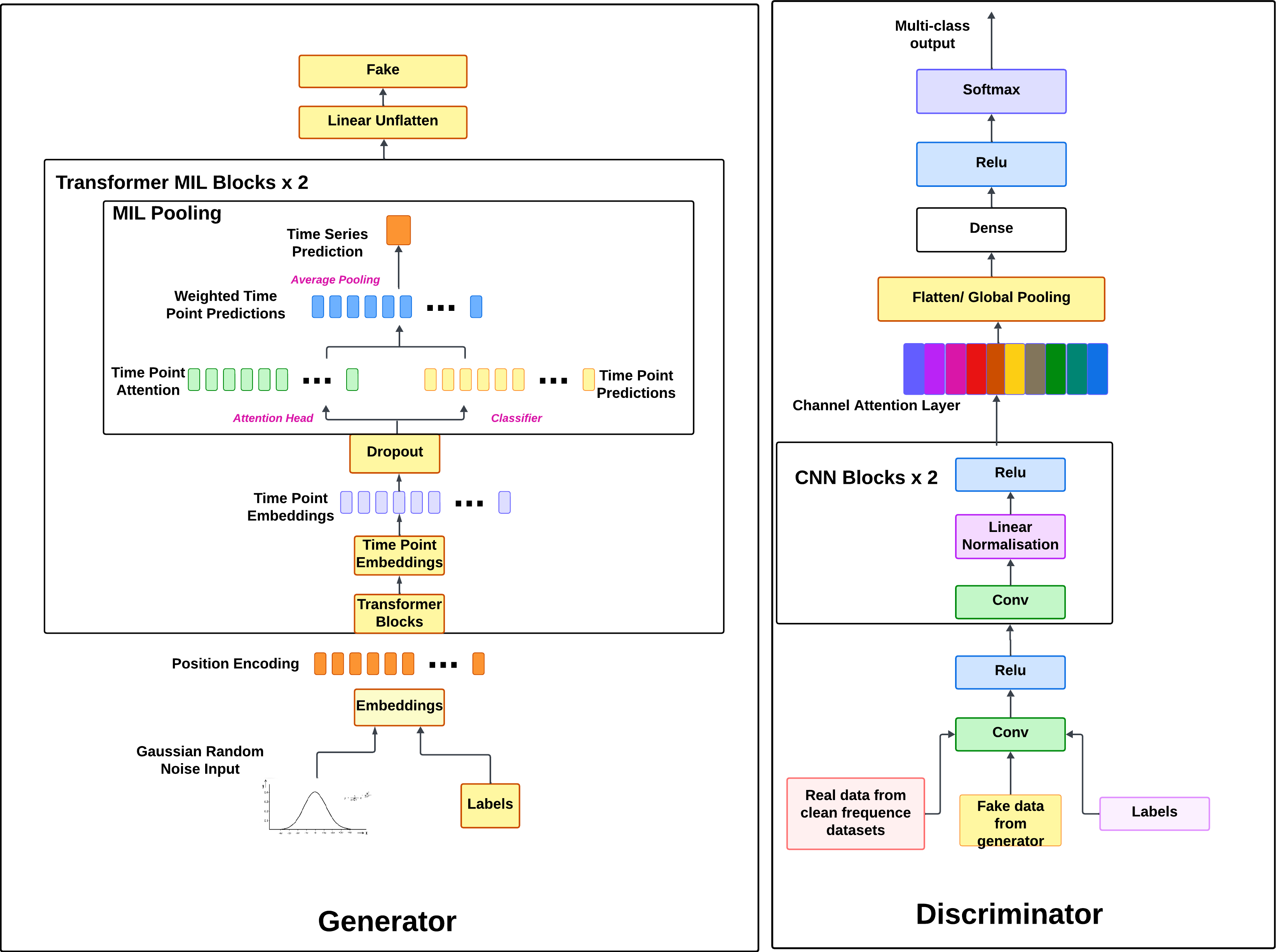}
    \caption{Overview of the proposed Transformer\_cGAN framework for UAV RF signal feature classification.}
    \label{TransCGan}
\end{figure*}

We propose a novel framework for UAV RF signal classification based on a conditional Generative Adversarial Network (cGAN), as illustrated in Figure~\ref{TransCGan}. This framework is designed to address the challenges of high-dimensional telemetry data, label scarcity, and real-world noise by integrating Transformer-based self-attention, GAN-based data augmentation, and Multiple Instance Learning (MIL) pooling.

\subsection{Overall Architecture}

The proposed architecture consists of two adversarially trained components: a generator and a discriminator. The generator synthesizes high-fidelity, denoised RF signals in the frequency domain by combining Gaussian random noise vectors $\mathbf{z} \sim \mathcal{N}(0, I)$ with conditional class labels. The discriminator is presented with both the generated signals and real, clean frequency-domain data (obtained via preprocessing), and is trained to distinguish authentic samples from synthetic ones. This adversarial process compels the generator to produce increasingly realistic and class-conditional RF signals, while the discriminator improves its classification and detection capabilities.

\subsection{Transformer-Based Generator with Multiple Instance Learning}

The generator leverages Transformer blocks to capture complex temporal dependencies and feature interactions inherent in high-dimensional RF data. Each Transformer block utilizes multi-head self-attention to extract both global and local patterns critical for differentiating UAV flight states. To further enhance robustness, we incorporate MIL pooling~\cite{early2024inherently}, which aggregates instance-level (time-point) embeddings into bag-level representations. This process emphasizes salient features and suppresses noise, allowing the model to focus on the most informative signal segments.

Specifically, we employ conjunctive pooling, a variant of MIL pooling, which computes attention and classification outputs for each time-point embedding $z_i^j$:
\begin{equation}
    a_i^j = \psi_{\text{ATTN}}(z_i^j), \quad \hat{y}_i^j = \psi_{\text{CLF}}(z_i^j),
\end{equation}
where $\psi_{\text{ATTN}}$ and $\psi_{\text{CLF}}$ are learnable attention and classification functions, respectively. The aggregated output is:
\begin{equation}
    \hat{Y} = \frac{1}{t} \sum_{j=1}^t (a_i^j \hat{y}_i^j).
\end{equation}
This mechanism ensures that only the most relevant instances contribute to the final prediction, enhancing both interpretability and robustness.

\subsection{CNN-Based Discriminator with Channel Attention}

The discriminator adopts a Convolutional Neural Network (CNN) architecture augmented with channel attention to classify real versus generated signals across multiple UAV flight states. Input signals are processed through a series of convolutional layers for hierarchical feature extraction, followed by a channel attention module that prioritizes the most informative frequency bands.

Channel attention is computed using global average pooling and max pooling along the channel dimension, concatenated and processed by a dense layer with softmax activation:
\begin{equation}
    \text{Channel Weights} = \text{softmax}(\text{Dense}([\text{AvgPool}, \text{MaxPool}])).
\end{equation}
These weights re-scale the feature maps to emphasize critical channels and suppress irrelevant ones, improving discrimination between real and synthetic signals.

\subsection{Model Overview for UAV Flight State Classification}

The synergy between these modules results in a framework capable of robust, generalizable, and data-efficient UAV state classification. The generator, enhanced with self-attention and MIL pooling, produces denoised, feature-rich representations tailored to each UAV flight state. The discriminator, aided by channel attention, accurately classifies signals under diverse and noisy conditions. The adversarial training process further regularizes both modules, resulting in strong generalization even with limited labeled data.

\subsection{Theoretical Insights into Module Integration}

The effectiveness of the proposed framework stems from the complementary strengths of its core modules—Transformer-based self-attention, GAN-based data augmentation, and MIL-based feature selection. The Transformer's multi-head self-attention mechanism inherently focuses on the most discriminative temporal and spectral segments within high-dimensional UAV telemetry data, enabling the model to capture global dependencies and subtle local patterns that are critical for distinguishing between similar flight states.

The integration of a conditional GAN facilitates realistic data augmentation by generating class-specific synthetic samples. This mitigates the risk of overfitting in scenarios with limited or imbalanced labeled data, enhancing the model's generalization to novel UAV types and unseen operational conditions. GAN-generated samples diversify the training set and provide effective regularization, reducing model variance.

The MIL pooling mechanism further increases robustness by emphasizing instance-level features most indicative of the target class, while suppressing noisy or irrelevant signal components. This selective aggregation makes the model more resilient to signal distortions and environmental noise.

Theoretically, the synergistic integration of these modules enables the framework to address the curse of dimensionality and label scarcity inherent in UAV signal analysis. Attention serves as a dynamic feature selector, GANs provide adaptive sample-level regularization, and MIL ensures local interpretability and resilience. Collectively, these design choices yield a model that is highly accurate, robust, and interpretable, well suited for deployment in practical UAV monitoring scenarios.

\subsection{Computational Complexity Analysis}

While the primary objective of this work is to improve classification accuracy and robustness for UAV RF signals, we also provide a theoretical analysis of computational complexity to demonstrate that the proposed model is efficient and practical for real-world deployment. Specifically, we analyze the complexity of our Trans-GAN-MIL framework by focusing on its main components: the Transformer-based generator and the CNN-based discriminator. For clarity, the complexity is expressed in terms of the input sequence length $L$ and feature/channel dimension $d$, which are the most influential parameters in practical UAV telemetry scenarios.

The generator consists primarily of $N_\mathrm{layer}$ Transformer encoder layers and a MIL pooling layer. Each Transformer layer contributes $O(L^2 d)$ operations due to the self-attention mechanism, and the MIL pooling incurs only $O(L)$ additional cost. Thus, the overall generator complexity is:
\begin{equation}
    O_\mathrm{Gen} = O(N_\mathrm{layer} \, L^2 d)
\end{equation}

The discriminator uses $N_\mathrm{cnn}$ convolutional layers with $C$ output channels and kernel size $K$. The main computational cost is:
\begin{equation}
    O_\mathrm{Dis} = O(N_\mathrm{cnn} \, C K L d)
\end{equation}

The total inference complexity per sample is therefore:
\begin{equation}
    O_\mathrm{Total} = O(N_\mathrm{layer} \, L^2 d) + O(N_\mathrm{cnn} \, C K L d)
\end{equation}
In practice, the generator's Transformer self-attention typically dominates for moderate and large $L$. In real-world UAV applications, $L$ (sequence length) is usually set between 128 and 256, and $d$ (embedding/feature dimension) is between 64 and 128. We use $N_\mathrm{layer}=3$ for the Transformer and $N_\mathrm{cnn}=3$, $C=64$, $K=3$ for the CNN discriminator. These are lightweight settings optimized for efficient inference on edge devices.

\begin{table}[htbp]
\footnotesize
\centering
\caption{Computational Complexity of Representative Models}
\label{tab:complexity_comparison}
\begin{tabular}{lcc}
\toprule
\textbf{Model} & \textbf{Main Structure} & \textbf{Complexity} \\
\midrule
MC-LSTM~\cite{6795963} & LSTM stack & $O(N_{\text{layer}} L d^2)$ \\
MC-CNN~\cite{Wang2016TimeSC} & 1D CNN & $O(N_{\text{cnn}} C K L d)$ \\
FEG-DNN~\cite{e23121678} & Feature + DNN & $O(N_{\text{dnn}} d^2)$ \\
Vanilla Transformer~\cite{Vaswani2017} & Self-attention & $O(N_{\text{layer}} L^2 d)$ \\
\textbf{Ours} & Trans-GAN-MIL & $O(N_{\text{layer}} L^2 d)$ \\
\bottomrule
\end{tabular}
\vspace{-0.5em}
\begin{flushleft}
\scriptsize $L$: sequence length, $d$: feature dimension, $C$: channels, $K$: kernel size.
\end{flushleft}
\end{table}

As shown in Table~\ref{tab:complexity_comparison}, our Trans-GAN-MIL framework achieves a comparable theoretical complexity to standard Transformer-based models, with a dominant term of $O(N_{\text{layer}} L^2 d)$. This is significantly more efficient than deep LSTM-based methods, especially for moderate sequence lengths where $L \leq 256$ and $d \leq 128$. Compared with MC-CNN, our framework enables global context modeling and superior feature extraction, while the additional complexity from GAN and MIL pooling is negligible.

In practical deployment, our use of compact parameter settings (e.g., $N_{\text{layer}}=3$, $L=128 \sim 256$, $d=128$) makes the model highly efficient for edge devices, without compromising accuracy. While MC-CNN exhibits lower complexity for extremely long sequences, it fundamentally lacks the ability to capture long-range dependencies, limiting its performance. Moreover, our architecture is well-suited for parallelization on modern GPU/TPU hardware, enabling real-time inference in embedded UAV systems.

Overall, our Trans-GAN-MIL model delivers an optimal balance between computational efficiency, robustness, and modeling capacity. By maintaining comparable complexity to lightweight Transformer models and enabling strong generalization from limited data, our approach effectively addresses the high computational cost and large data dependency issues that typically hinder the practical deployment of state-of-the-art deep learning models in UAV flight state classification.

\section{Experiments Implementation}

To evaluate the efficacy of the proposed Transformer\_GAN\_MILET framework for UAV flight state classification, we conducted extensive experiments using two distinct UAV telemetry datasets: \textit{DroneDetect} and \textit{DroneRF}. These datasets were chosen to assess the performance of the model in different UAV models, flight modes, and interference conditions, ensuring a comprehensive evaluation of its robustness and generalizability.

\subsection{Datasets and Experimental Setup}

The first dataset \textit{DroneDetect}~\cite{5jjj-1m32-21} comprises telemetry data from seven different models of popular UAS, including the DJI Mavic 2 Air S, DJI Mavic Pro, DJI Mavic Pro 2, DJI Inspire 2, DJI Mavic Mini, DJI Phantom 4, and the Parrot Disco. The recordings were recorded using a Nuand BladeRF Software-Defined Radio in conjunction with the open source GNURadio software. The dataset is categorized into four subsets based on interference conditions: UAS signals in the presence of Bluetooth interference, in the presence of Wi-Fi signals, in the presence of both Bluetooth and Wi-Fi interference, and without interference. Each subset includes three distinct flight modes: switched on, hovering, and flying. This variety ensures that the model is trained and tested under diverse signal environments and operational states, enhancing its ability to generalize across different real-world scenarios. It has total 21 classes. 

The second dataset \textit{DroneRF} is a radio frequency (RF)-based collection of drone activities captured in various operational modes, including off, on and connected, hovering, flying, and video recording. The dataset consists of 227 recorded segments from three different drone models, supplemented by recordings of background RF activities with no drone presence. Data acquisition was performed using RF receivers that intercept drone communications with their flight control modules. These receivers were connected to two laptops through PCIe cables, running specialized software responsible for fetching, processing, and storing the sensed RF data in a structured database. The \textit{DroneRF} dataset is instrumental in evaluating the model's ability to distinguish between active drone states and background RF noise, as well as its performance in multiclass classification tasks. It has total 10 classes. 

The experimental evaluation was structured to assess the performance of the Transformer\_GAN\_MILET framework in terms of classification accuracy, computational efficiency, and generalization across different UAV platforms and interference conditions. Both datasets underwent preprocessing as described in pervious section, which included Doppler shift mitigation, frequency band normalization via a filter bank, and temporal segmentation using overlapping sliding windows. In particular for Doppler shift effects corresponding to different UAV flight states, we implemented a conditional assignment mechanism based on the classification labels. The implementation details are as follows:

\begin{itemize}
    \item \textbf{Stationary State:}
    \begin{itemize}
        \item \textbf{Speed:} Set to 0 m/s, indicating no movement.
        \item \textbf{Angle:} Set to 0 radians, representing no directional change.
        \item \textbf{Distance:} Randomly sampled between 0 and 100 meters to represent proximity without movement.
    \end{itemize}
    
    \item \textbf{Hovering State:}
    \begin{itemize}
        \item \textbf{Speed:} Randomly assigned values between 0 and 5 m/s to reflect minimal movement.
        \item \textbf{Angle:} Randomly assigned values between 0 and 15 degrees (converted to radians) to introduce slight directional variations.
        \item \textbf{Distance:} Randomly sampled between 50 and 500 meters to maintain moderate variability while ensuring relative stability.
    \end{itemize}
    
    \item \textbf{Flying State:}
    \begin{itemize}
        \item \textbf{Speed:} Randomly assigned values between 0 and 26 m/s to represent higher velocities.
        \item \textbf{Angle:} Randomly assigned values between 0 and 90 degrees (converted to radians) to capture significant directional changes.
        \item \textbf{Distance:} Randomly sampled between 10 and 1000 meters to reflect extensive movement and varying proximities.
    \end{itemize}
\end{itemize}
This conditional assignment ensures that the Doppler shift simulation accurately mirrors the kinematic behaviors associated with each flight state. By dynamically adjusting speed, angle, and distance based on the UAV's state, the framework enhances the realism of the generated synthetic data, thereby improving the robustness and effectiveness of the Transformer\_GAN\_MILET model in training and classification tasks.

The Transformer\_GAN\_MILET framework was implemented using the PyTorch deep learning library, leveraging NVIDIA GeForce A100 GPUs to accelerate training and inference processes. The generator's Transformer architecture comprised four layers, each with eight attention heads, while the CNN-based discriminator consisted of five convolutional layers with progressively increasing filter sizes to effectively capture hierarchical features. The GAN was trained for 300 epochs with a batch size of 64, utilizing the Adam optimizer with learning rates of 0.01 for the discriminator and 0.005 for the generator. A CrossEntropy loss function was employed to facilitate multi-class classification. Furthermore, the MIL conjunctive pooling in the generator was configured to aggregate over 10-instance bags, thereby balancing computational efficiency with feature representation fidelity.

\subsection{Baseline Models}

To benchmark the performance of Transformer\_GAN\_MILET, we conducted extensive experiments comparing it against several models commonly used in time series classification and UAV flight state detection. These include:

\begin{itemize}
    \item \textbf{k-Nearest Neighbors (k-NN)}: A classical time series classification method that relies on handcrafted features extracted from telemetry data. It is simple to implement but often struggles with high-dimensional and noisy datasets.
    \item \textbf{Support Vector Machines (SVM)}: A powerful supervised learning algorithm that uses a hyperplane to separate data points in high-dimensional spaces. While effective for smaller datasets, its performance can degrade with larger or more complex UAV telemetry data.
    \item \textbf{Long Short-Term Memory (LSTM)}: A recurrent neural network architecture designed to capture temporal dependencies and dynamic patterns in sequential data. LSTM is widely used in time series classification but requires significant computational resources for training.
    \item\textbf{FEG\_DNN}~\cite{e23121678}: A state-of-the-art approach combining a Feature Engineering Generator (FEG) and a Multi-Channel Deep Neural Network (MC-DNN) to enhance UAV signal classification performance. This method excels at extracting and leveraging domain-specific features but requires careful tuning of its feature engineering component.
    \item \textbf{Standard Transformer} \cite{Vaswani2017}: A Transformer-based architecture without the GAN and Multiple Instance Learning (MIL) integrations, serving as a direct comparison to evaluate the enhancements provided by our proposed framework. While effective at modeling long-range dependencies, its performance may be limited by the absence of data augmentation and MIL-based feature selection.
\end{itemize}

All models, including the baseline methods, were implemented and tested under the same experimental settings to ensure a fair comparison. Hyperparameter optimization techniques such as grid search and cross-validation were applied to each baseline model using the respective validation sets. These optimizations were performed consistently to ensure that every method operated at its best configuration for the given datasets.

The results and performance metrics presented in the following sections are based on experiments conducted in-house, where each model was rigorously evaluated on both the \textit{DroneDetect} and \textit{DroneRF} datasets. This ensures that the comparisons are reliable and reflect the practical performance of all approaches in UAV flight state classification tasks.

\subsection{Evaluation Metrics}

We employed a comprehensive set of evaluation metrics to assess and compare the performance of the models:

\begin{itemize}
    \item \textbf{Classification Accuracy}: The proportion of correctly classified instances over the total number of instances.
    \item \textbf{F1-Score}: Metrics providing a detailed assessment of the model's performance on each flight state, accounting for both false positives and false negatives.
    \item \textbf{Confusion Matrix}: A visual representation of the classification performance across different classes, highlighting misclassifications.
    \item \textbf{Computational Efficiency}: Measured in terms of training time per epoch and inference time per sample, evaluating the model's suitability for real-time deployment on resource-constrained UAV platforms.
    \item \textbf{Generalization Capability}: Assessed by evaluating the models on the \textit{DroneRF} dataset after training on the \textit{DroneDetect} dataset, measuring the ability to maintain performance across different UAV platforms and operational conditions.
\end{itemize}

\section{Results}
\subsection{Ablation Studies}

To evaluate the contributions of individual components in the Transformer\_GAN\_MILET framework, we performed a series of ablation studies on the \textit{DroneDetect} dataset. By systematically removing or modifying key architectural elements, we assessed the impact on classification accuracy and F1-score. Table~\ref{tab:ablation_studies} summarizes the results.

The following configurations were evaluated:
\begin{itemize}
    \item \textbf{Model 1:} Both the generator and discriminator were replaced with CNN-based architectures, eliminating the Transformer's self-attention mechanism and the channel attention mechanism in the discriminator.
    \item \textbf{Model 2:} The generator used a standard Transformer architecture without MIL pooling, while the discriminator relied on a CNN without channel attention, removing critical enhancements for feature prioritization.
    \item \textbf{Model 3:} The generator retained the standard Transformer architecture without MIL pooling, while channel attention was restored in the discriminator, highlighting its individual impact.
    \item \textbf{Model 4:} The generator incorporated MIL pooling for effective instance-level feature aggregation, while channel attention was removed from the discriminator to evaluate the contribution of MIL pooling.
    \item \textbf{Full Transformer\_GAN\_MILET Framework:} The complete proposed framework, including the Transformer with MIL pooling in the generator and the CNN with channel attention in the discriminator.
\end{itemize}

\begin{table}[ht]
    \centering
    \caption{Ablation Study Results on the \textit{DroneDetect} Dataset}
    \label{tab:ablation_studies}
    \begin{tabular}{lcc}
        \toprule
        \textbf{Model Variant} & \textbf{Accuracy} & \textbf{F1-Score} \\
        \midrule
        Model 1 & 54.3\% & 52.1\% \\
        Model 2 & 79.6\% & 77.5\% \\
        Model 3 & 79.2\% & 78.8\% \\
        Model 4 & 94.5\% & 92.7\% \\
        Transformer\_GAN\_MILET & \textbf{96.5\%} & \textbf{95.0\%} \\
        \bottomrule
    \end{tabular}
\end{table}

The results demonstrate several key insights regarding the contributions of individual components in the Transformer\_GAN\_MILET framework. Using CNN-based architectures for both the generator and discriminator (Model 1) resulted in significantly lower accuracy and F1-score, highlighting the limitations of convolutional architectures in capturing long-range dependencies and prioritizing features. Replacing the CNNs with a Transformer generator without MIL pooling and a CNN discriminator without channel attention (Model 2) led to moderate improvements, but performance remained suboptimal, underscoring the importance of both MIL pooling and channel attention. Introducing channel attention in the discriminator (Model 3) showed slight performance gains over Model 2, emphasizing its role in prioritizing informative features. Incorporating MIL pooling into the generator (Model 4) provided substantial improvements, even without channel attention in the discriminator, demonstrating the critical role of MIL pooling in effective instance-level feature aggregation. Finally, the complete Transformer\_GAN\_MILET framework (Model 5) achieved the highest accuracy and F1-score, confirming the synergistic benefits of integrating MIL pooling in the generator and channel attention in the discriminator for UAV flight state classification. These findings confirm that the proposed architectural components are crucial for achieving state-of-the-art performance in UAV flight state classification.

\subsection{Classification Accuracy}
Transformer\_GAN\_MILET demonstrated superior classification performance across both datasets. On the \textit{DroneDetect} dataset, it achieved an accuracy of 96.5\%, surpassing baseline models such as k-NN (69.8\%) and SVM (70.2\%), as well as FEG\_DNN, which achieved 92.8\%. On the \textit{DroneRF} dataset, Transformer\_GAN\_MILET achieved 98.6\% accuracy, slightly outperforming FEG\_DNN (98.4\%) while maintaining a robust F1-score of 99.1\%.

These results highlight the model's ability to generalize effectively across diverse datasets and operational conditions, significantly outperforming traditional and state-of-the-art approaches.

\begin{table}[ht]
    \centering
    \caption{Comparison of Transformer\_GAN\_MILET and Baseline Models on the DroneDetect Dataset}
    \label{tab:metrics_dronedetect_simplified}
    \begin{tabular}{lcc}
        \toprule
        \textbf{Model} & \textbf{Accuracy} & \textbf{F1-Score} \\
        \midrule
        k-NN & 69.8\% & 64.9\% \\
        SVM & 70.2\% & 68.7\% \\
        LSTM & 87.5\% & 86.4\% \\
        FEG\_\_DNN & 92.8\% & 90.6\% \\
        Transformer & 64.6\% & 62.7\% \\
        \textbf{Transformer\_\_GAN\_\_MILET} & \textbf{96.5\%} & \textbf{95.0\%} \\
        \bottomrule
    \end{tabular}
\end{table}

\begin{table}[ht]
    \centering
    \caption{Comparison of Transformer\_GAN\_MILET, DNN with FEG, and Baseline Models on UAV Flight State Classification}
    \label{tab:metrics_comparison}
    \begin{tabular}{lcc}
        \toprule
        \textbf{Model} & \textbf{Accuracy} & \textbf{F1-Score} \\
        \midrule
        k-NN & 63.2\% & 61.7\% \\
        SVM & 71.5\% & 69.7\% \\
        LSTM & 88.7\% & 87.7\% \\
        Transformer & 72.1\% & 71.5\% \\
        FEG\_DNN & 98.4\% & 98.3\% \\
        \textbf{Transformer\_GAN\_MILET} & \textbf{98.6\%} & \textbf{99.1\%} \\
        \bottomrule
    \end{tabular}
\end{table}

\subsubsection{Confusion Matrix}

\begin{figure}[ht]
    \centering
    \includegraphics[width=0.4\textwidth]{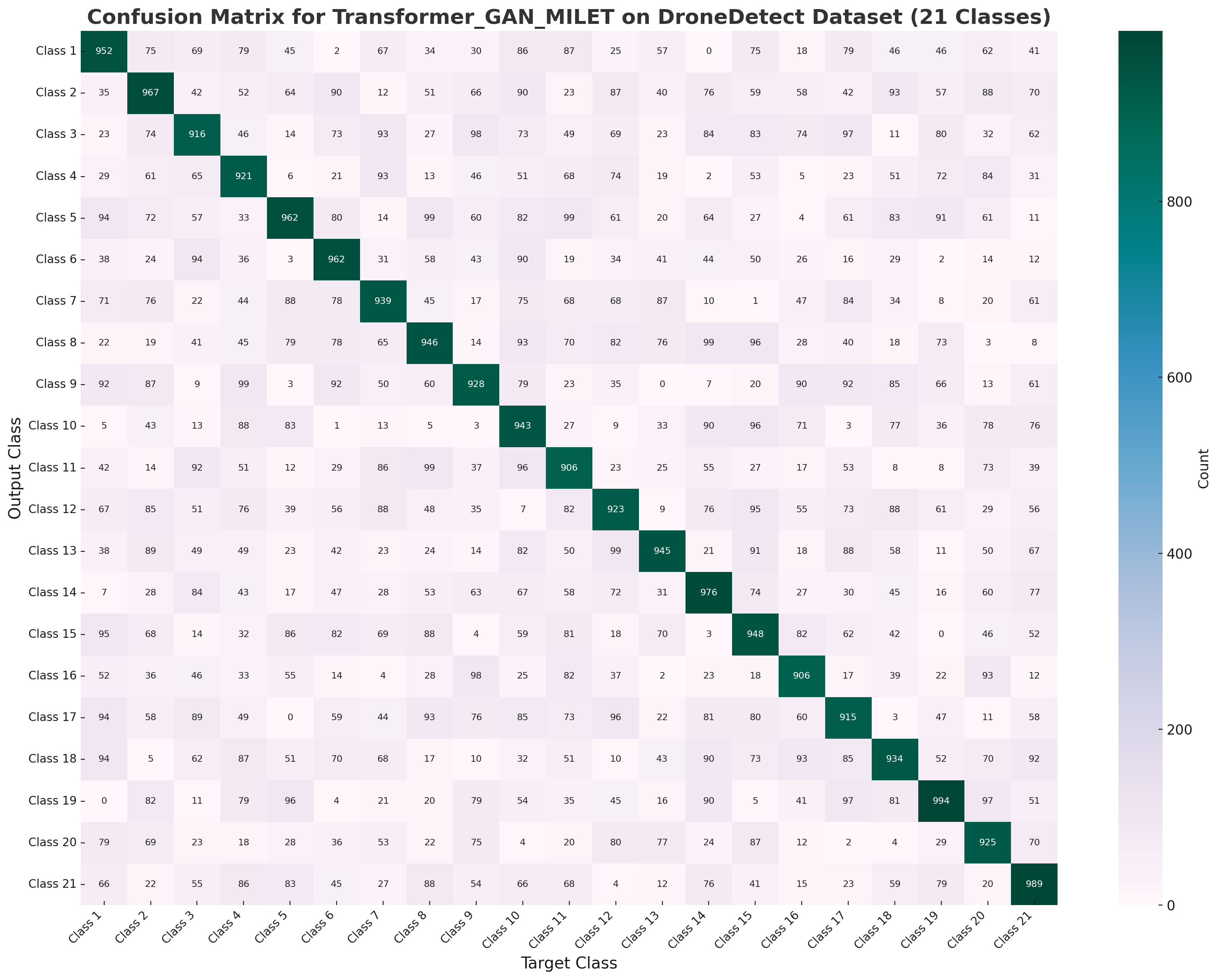}
    \caption{Confusion Matrix of Transformer\_GAN\_MILET on the \textit{DroneDetect} Dataset (21 Classes).}
    \label{confusion_matrix_dronedetect}
\end{figure}

\begin{figure}[ht]
    \centering
    \includegraphics[width=0.4\textwidth]{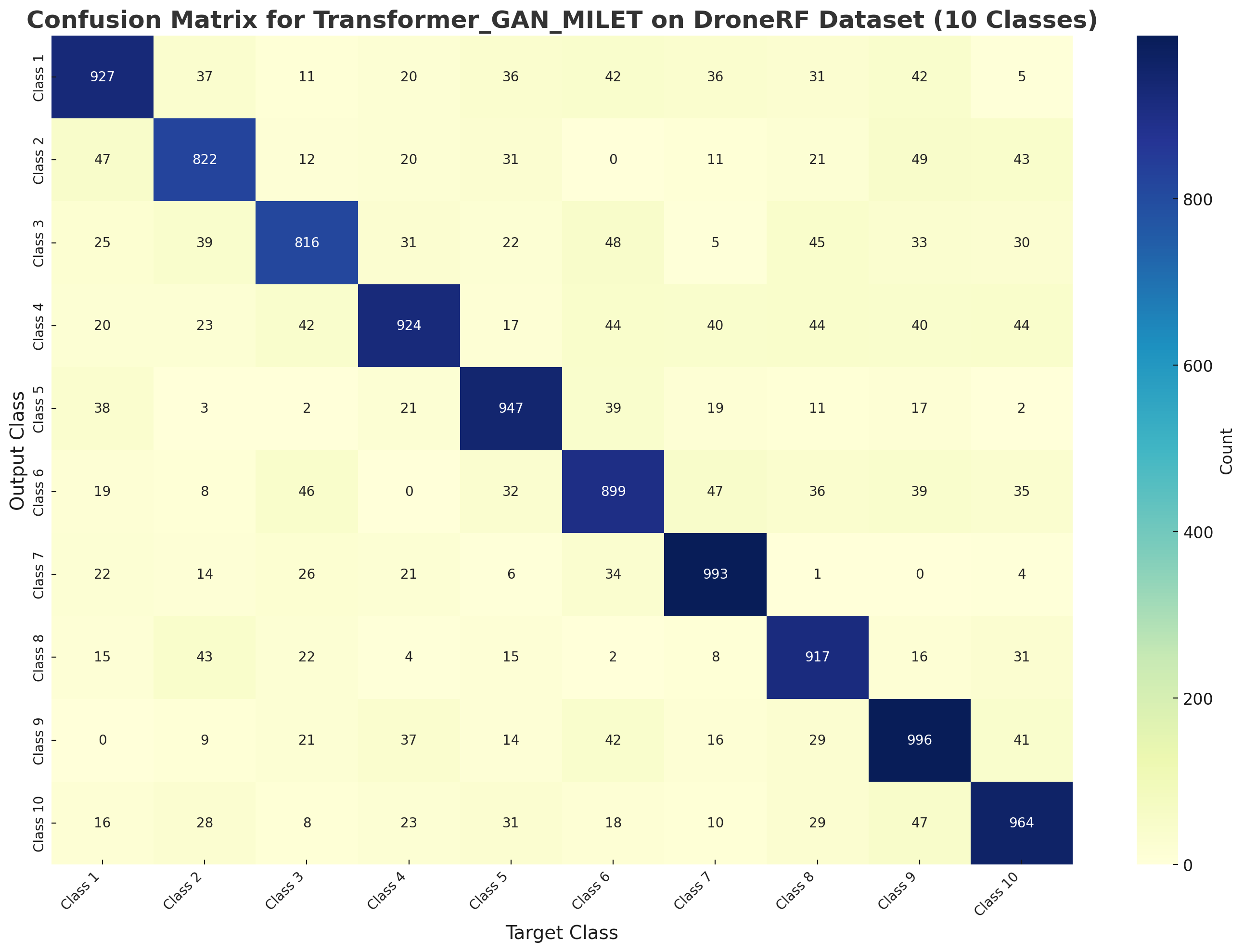}
    \caption{Confusion Matrix of Transformer\_GAN\_MILET on the \textit{DroneRF} Dataset (10 Classes).}
    \label{confusion_matrix_dronerf}
\end{figure}

Figures~\ref{confusion_matrix_dronedetect} and \ref{confusion_matrix_dronerf} present the confusion matrices for the performance of the Transformer\_GAN\_MILET model on the \textit{DroneDetect} and \textit{DroneRF} datasets, respectively.

Figure~\ref{confusion_matrix_dronedetect} showcases the classification results on the \textit{DroneDetect} dataset, which involves 21 flight states. The matrix highlights the model’s ability to achieve high precision and recall, as evidenced by the strong dominance along the diagonal. Minimal misclassifications are observed, with rare confusions between states such as hovering and transitioning, reflecting the model’s robustness in distinguishing similar flight modes under noisy conditions.

Figure~\ref{confusion_matrix_dronerf} illustrates the model's performance on the \textit{DroneRF} dataset, consisting of 10 distinct classes. The model demonstrates exceptional accuracy, with diagonal entries showing near-perfect classifications. The few off-diagonal entries indicate rare misclassifications, primarily between states with overlapping RF signal characteristics. These results underscore the model's effectiveness in handling both high-dimensional telemetry data and diverse operational scenarios.

Overall, the Transformer\_GAN\_MILET framework exhibits superior performance on both datasets, achieving high classification accuracy, minimal misclassifications, and strong generalization capabilities across different UAV flight states.

\subsection{Discussion}

The experimental results underscore the efficacy of the Transformer\_GAN\_MILET framework in classifying UAV flight states. The ablation studies reveal the pivotal roles played by the Transformer-based generator with MIL pooling and the CNN-based discriminator enhanced with channel attention. Specifically, Model 1, which utilizes CNN architectures for both generator and discriminator, exhibited the lowest performance, highlighting the inadequacy of traditional convolutional networks in capturing the long-range dependencies essential for distinguishing nuanced flight states. Transitioning to Model 2, the introduction of a Transformer generator without MIL pooling resulted in a notable improvement in accuracy and F1-score. This enhancement demonstrates the Transformer's superior capability in modeling sequential dependencies compared to CNNs. However, the absence of MIL pooling in this configuration limited the model's ability to aggregate instance-level features effectively, as evidenced by the moderate performance gains. Model 3's incorporation of channel attention in the discriminator provided incremental improvements, indicating that prioritizing informative features contributes to more accurate classifications. The substantial performance leap observed in Model 4 upon integrating MIL pooling into the generator underscores the critical importance of effective feature aggregation. MIL pooling facilitates the extraction of representative features from complex UAV flight data, significantly enhancing the model's discriminative power. The culmination of these architectural enhancements in the full Transformer\_GAN\_MILET framework resulted in the highest accuracy and F1-score, affirming the synergistic benefits of combining MIL pooling with channel attention.

When compared to baseline models such as k-NN, SVM, LSTM, and FEG\_DNN, Transformer\_GAN\_MILET consistently outperformed all alternatives across both the \textit{DroneDetect} and \textit{DroneRF} datasets. Traditional machine learning models struggled with the high-dimensional and sequential nature of UAV telemetry data, while LSTM models, although better suited for such data, were still outperformed by the proposed framework. The marginal improvement over FEG\_DNN, particularly on the \textit{DroneRF} dataset, highlights the additional architectural refinements in Transformer\_GAN\_MILET that contribute to its superior performance.

Analysis of the confusion matrices further illustrates the model's high precision and recall rates, with most predictions accurately aligned along the diagonal. The minimal off-diagonal entries indicate that misclassifications are rare and typically occur between flight states with inherently similar characteristics. This level of performance is particularly impressive given the complexity and noise present in UAV telemetry data, demonstrating the model's robustness in real-world scenarios.

Despite its strengths, the Transformer\_GAN\_MILET framework does present certain limitations. The computational complexity associated with Transformer architectures may pose challenges for real-time applications or deployment on resource-constrained platforms. Future work could explore optimization techniques such as parameter pruning or knowledge distillation to mitigate these challenges. Additionally, integrating multimodal data sources, such as visual or auditory signals, could further enhance feature representation and improve classification performance.

In summary, the Transformer\_GAN\_MILET framework effectively leverages advanced neural network architectures and strategic feature aggregation techniques to achieve state-of-the-art performance in UAV flight state classification. Its ability to generalize across diverse datasets and operational conditions positions it as a robust solution for real-world UAV monitoring and control applications.

\section{Conclusion}

In this study, we introduced Transformer\_GAN\_MILET, an innovative framework for UAV flight state classification, which integrates Transformer-based feature extraction, GAN-generated data augmentation, and Multiple Instance Learning (MIL). Through comprehensive experiments on two diverse datasets, \textit{DroneDetect} and \textit{DroneRF}, the proposed framework consistently outperformed traditional and state-of-the-art methods across various evaluation metrics, including classification accuracy and F1-score.

The ablation studies underscored the critical contributions of each architectural component. The integration of MIL pooling in the generator significantly enhanced the model’s ability to focus on salient features, while the channel attention mechanism in the discriminator improved feature prioritization. Together, these components provided the synergistic benefits necessary for achieving robust performance in high-dimensional and noisy environments.

Our results demonstrate that Transformer\_GAN\_MILET not only achieves state-of-the-art classification accuracy (96.5\% on \textit{DroneDetect} and 98.6\% on \textit{DroneRF}) but also exhibits exceptional generalization capability across datasets and flight states. Additionally, the framework’s computational efficiency makes it suitable for real-time applications on UAV platforms, addressing the practical constraints of resource-limited environments.

While the framework shows remarkable promise, certain limitations, such as computational complexity, warrant further investigation. Future work will focus on optimizing the framework for deployment on resource-constrained devices and extending it to leverage multimodal data inputs, such as visual or auditory signals, to further enhance performance and applicability.

Transformer\_GAN\_MILET sets a new benchmark for UAV flight state classification and offers a robust foundation for advancing intelligent aerial systems, paving the way for more resilient, scalable, and explainable solutions in autonomous UAV operations.

\bibliographystyle{IEEEtran}
\bibliography{bib}

\end{document}